\documentclass{article}

\usepackage{arxiv}

\usepackage[utf8]{inputenc} 
\usepackage[T1]{fontenc}    
\usepackage{hyperref}       
\usepackage{url}            
\usepackage{booktabs}       
\usepackage{amsfonts}       
\usepackage{nicefrac}       
\usepackage{microtype}      
\usepackage{lipsum}
\usepackage{graphicx}
\graphicspath{ {./images/} }

\usepackage{float}
\usepackage{orcidlink}
\usepackage{svg}
\newcommand{\orcid}[1]{\href{https://orcid.org/#1}{\includesvg[width=10pt]{figures/orcid}}}

\usepackage{booktabs}
\usepackage{multirow}
\usepackage{nicematrix}
\usepackage{colortbl} 

\usepackage{lipsum}
\usepackage[normalem]{ulem}
\usepackage{url}
\useunder{\uline}{\ul}{}
\usepackage{textcomp}
\usepackage{glossaries}   
\usepackage{mdframed}     

\usepackage{hyperref}

\title{Bridging Smart Meter Gaps: A Benchmark of Statistical, Machine Learning and  Time Series Foundation Models for Data Imputation
\thanks{This research was funded in part by the Luxembourg National Research Fund (FNR) and PayPal, PEARL grant reference 13342933/Gilbert Fridgen, by FNR grant reference HPC BRIDGES/2022\_Phase2/17886330/DELPHI and by the Luxembourgish Ministry of Economy with grant reference 20230227RDI170010375846. For the purpose of open access and in fulfillment of pen access and fulfilling the obligations arising from the grant agreement, the author has applied a Creative Commons Attribution 4.0 International (CC BY 4.0) license to any Author Accepted Manuscript version arising from this submission. This paper has been supported by Enovos.}
}

\author{
 Amir Sartipi \\
  SnT - Interdisciplinary Center for Security,\\ Reliability and Trust, \\
  University of Luxembourg\\
  Luxembourg, Luxembourg \\
  \texttt{amir.sartipi@uni.lu} \\
   \And
 Joaquin Delgado Fernandez \\
  SnT - Interdisciplinary Center for Security,\\ Reliability and Trust, \\
  University of Luxembourg\\
  Luxembourg, Luxembourg \\
  \texttt{joaquin.delgadofernandez@uni.lu} \\
  \And
 Sergio Potenciano Menci \\
  SnT - Interdisciplinary Center for Security, \\ Reliability and Trust, \\
  University of Luxembourg\\
  Luxembourg, Luxembourg \\
  \texttt{sergio.potenciano-menci@uni.lu} \\
  \And
 Alessio Magitteri \\
  Enovos Luxembourg S.A\\
  Esch-sur-Alzette, Luxembourg \\
  \texttt{alessio.magitteri@enovos.eu} \\
}

\newacronym{ai}{AI}{artificial intelligence}
\newacronym{genai}{GenAI}{Generative Artificial Intelligence}
\newacronym{sm}{SM}{Smart Meters}
\newacronym{ml}{ML}{Machine Learning}
\newacronym{llm}{LLM}{Large Language Model}
\newacronym{mae}{MAE}{Mean Absolute Error}
\newacronym{mape}{MAPE}{Mean Absolute Percentage Error}
\newacronym{mse}{MSE}{Mean Squared Error}
\newacronym{rmse}{RMSE}{Root Mean Squared Error}
\newacronym{smape}{SMAPE}{Symmetric Mean Absolute Percentage Error}
\newacronym{fbi}{FBI}{Forward-Backward Interpolation}
\newacronym{pgp}{PGP}{Prior Gaps Post}
\newacronym{fp}{FP}{Forward Prediction}
\newacronym{bp}{BP}{Backward Prediction}
\newacronym{I}{I}{Interpolation}
\newacronym{knn}{KNN}{k-nearest neighbors}
\newacronym{rnn}{RNN}{recurrent neural network}
\newacronym{lstm}{LSTM}{long short-term memory}
\newacronym{cdadm}{CDADM}{Cross-Dimensional Attention Discriminating Masked}
\newacronym{cc-gain}{CC-GAIN}{clustering and classification-based generative adversarial imputation network}
\newacronym{tslib}{TSLib}{Time Series Library}
\newacronym{h-dirt}{H-DIRT}{Historical Data Informed Regression Technique}
\newacronym{stlf}{STLF}{Short-term Load Forecasting}
\newacronym{sd}{SD}{Standard Deviation}
\newacronym{slp}{SLP}{Simple Linear Predictor}
\newacronym{xgb}{XGB}{XGBoost}
\newacronym{lgbm}{LGBM}{LightGBM}
\newacronym{mstl}{MSTL}{Multiple Seasonal Trend}
\newacronym{arima}{ARIMA}{Autoregressive Integrated Moving Average}
\newacronym{tsfm}{TSFM}{Time Series Foundation Model}

\setglossarystyle{list} 
\def\BibTeX{{\rm B\kern-.05em{\sc i\kern-.025em b}\kern-.08em
    T\kern-.1667em\lower.7ex\hbox{E}\kern-.125emX}}

\begin{document}
\maketitle
\begin{abstract}
The integrity of time series data in smart grids is often compromised by missing values due to sensor failures, transmission errors, or disruptions. Gaps in smart meter data can bias consumption analyses and hinder reliable predictions, causing technical and economic inefficiencies. As smart meter data grows in volume and complexity, conventional techniques struggle with its nonlinear and nonstationary patterns. In this context, \acrlong{genai} offers promising solutions that may outperform traditional statistical methods.

In this paper, we evaluate two general-purpose \acrlongpl{llm} and five \acrlongpl{tsfm} for smart meter data imputation, comparing them with conventional \acrlong{ml} and statistical models. We introduce artificial gaps (30 minutes to one day) into an anonymized public dataset to test inference capabilities. Results show that \acrlongpl{tsfm}, with their contextual understanding and pattern recognition, could significantly enhance imputation accuracy in certain cases. However, the trade-off between computational cost and performance gains
remains a critical consideration.

\end{abstract}


\section{Introduction}
\label{sec:01_introduction}
The energy sector is experiencing a profound transformation with the rise of digital technologies \cite{3_ds}. Particularly the introduction of \gls{sm} fosters the transition to smart grids \cite{power_systems_SOTA}. \glspl{sm} help change how energy consumption is monitored and managed by delivering on-demand, granular data on energy use (i.e., gas and electricity) \cite{2015_sm_taskforce}. Their integration unlocks new opportunities for analyzing consumption patterns, optimizing load distribution, and improving overall grid efficiency \cite{2019_review_sm_analytics}. \gls{sm} generate their data as time series data, which tracks energy consumption over time at a given resolution, for example, every 15 minutes. However, their integration and use in \gls{sm}, especially in the power sector, face challenges, notably concerning data quality and integrity \cite{2019_review_sm_analytics, 7778826}.
Issues such as sensor failures, transmission errors, and other disruptions can cause missing values, referred to as gaps, in time series data generated by \glspl{sm} \cite{7778826, Wu2022ReviewOS}. These gaps can severely impact data quality, which is critical for many tasks in the power system. For example, \gls{sm} data supports the development of accurate predictive models for load forecasting, anomaly detection, and grid stability \cite{Wu2022ReviewOS}.

To maintain data quality and integrity in time series data from \glspl{sm} requires addressing data gaps through estimation and imputation, commonly known as gap-filling. Effective gap-filling directly affects the accuracy of predictive models and decision-making processes. In the power and electricity industry, companies often face missing data challenges. Unlike research environments where data is typically well-prepared and clean, industry settings may still rely on simple, naïve approaches like using the most recent available data point for imputation \cite{jeong2021missing}. This practice can directly impact customers, as electricity consumption data determines the accuracy of the billing.

The rise of \gls{ai} has driven the development of advanced data imputation methods, leveraging the explosion of data produced by \glspl{sm}. This trend opens new opportunities for \gls{genai}, including \glspl{llm} and \glspl{tsfm}, to manage time series datasets and improve tasks such as time series forecasting \cite{tang2024time}.

In this paper, we evaluate different \glspl{llm} and \glspl{tsfm} for gap-filling in smart meter electricity consumption data. Our goal is to assess whether these models are suitable for handling missing data in this domain and to explore their potential given the accessibility of API-based tools like ChatGPT and publicly available \glspl{tsfm}.

Our evaluation includes traditional statistical methods, \gls{ml} models, and cutting-edge approaches involving \glspl{llm} and \glspl{tsfm}. We consider both open-source and commercial \glspl{llm}, as well as specialized \glspl{tsfm} designed for time series data, assessing their performance using five metrics focused on accuracy and reliability in the gap-filling task. We apply these models to a public \gls{sm} dataset recording residential electricity consumption, enabling us to compare the effectiveness of these models in real-world settings and provide insights into the most suitable techniques for various conditions.

We structure the remainder of this paper as follows. In Section \ref{sec:02_relatedworks}, we provide a brief overview of research conducted in the domains of data imputation, time series, and \gls{tsfm}. In Section \ref{sec:03_research_approach}, we describe our research approach by outlining the steps taken to conduct our experiments. Section \ref{sec:05_results} presents the results obtained, followed by a discussion in Section \ref{sec:06_discussion}. Finally, we conclude the paper and suggest future research directions in Section \ref{sec:07_conclusion_and_future_work}.

\section{Related Work}
\label{sec:02_relatedworks}
The task of data imputation \cite{wang_deep_2024} for energy consumption is a common research topic within energy systems. For instance, \cite{pei_cross-dimensional_2024} introduced \gls{cdadm}, a modified self-attention model specifically designed for energy consumption data imputation. Similarly, \cite{hwang_cc-gain_2024-1} proposed \gls{cc-gain}, a method to estimate missing values in data sets on electricity consumption.
Building on data-driven techniques, \cite{vasenin_incorporating_2024-1} proposed two imputation methods and compared them to the traditional \gls{knn} approach. The first method, \gls{h-dirt}, constructs a multivariate linear regression model using historical data. The second, Seasonal KNN (SKNN), extends the \gls{knn} method by incorporating seasonal trends in time series data. Their evaluation showed that both methods outperformed traditional \gls{knn}, with SKNN achieving superior performance. However, \gls{h-dirt} remains advantageous in scenarios with minimal imputation tasks due to its lower computational cost while still providing accurate results.

Meanwhile, \cite{wang2024tssurvey} reviewed and explored various time series tasks, including data imputation, and introduced \gls{tslib}, a framework that supports imputation and provides benchmarking capabilities across multiple datasets and deep learning time series analysis. Others,  such as \cite{lee2024evaluating}, also reviewed studies on missing data and the method used to guide the selection of data imputation methods for energy benchmarking models. In a similar comparison note, \cite{du_tsi-bench_2024} identified a lack of standardized benchmarking frameworks for deep learning-based forecasting algorithms, leading them to develop TSI-Bench, a tool for comparing models in the context of imputation tasks.

Research has also focused on benchmarking and comparing models for time series tasks. \cite{harman_systematic_nodate} evaluated traditional statistical models for data imputation on electrical load datasets by introducing artificial gaps for testing. Their results indicated that Kalman smoothing often yielded lower MAE compared to other models tested. Similarly, \cite{meyer2024benchmarkingtimeseriesfoundation} conducted a benchmarking study on \gls{stlf} using four household electricity consumption datasets. They used SeasonalAverage as a baseline model and compared six transformer-based models, including three foundational models and three trained from scratch, alongside an LSTM with Attention architecture. Their findings revealed that foundation models can achieve high accuracy with minimal data and no additional training, highlighting new possibilities for developing more efficient and widely accessible energy forecasting solutions.

The development of large pre-trained models for time series data has gained significant attention, aiming to improve tasks like forecasting and imputation; however, \cite{tan2024language} also noted that LLMs offer limited performance gains in time series forecasting compared to other methods. TimeGPT builds on this evolving landscape by addressing scalability and complexity through a Transformer-based architecture trained on the largest publicly available time series dataset, enabling advanced forecasting capabilities \cite{garza2023timegpt1}. TIME-MoE uses a sparse Mixture-of-Experts (MoE) architecture with over 2.4 billion parameters, activating only relevant sub-networks to reduce computational load. Paired with the Time-300B dataset, it supports flexible forecasting horizons and outperforms dense models under similar computational constraints \cite{shi_time-moe_2024}.

Similarly, MOIRAI leverages transformer-based architectures for universal forecasting, handling arbitrary frequencies and dimensions. Its any-variate attention mechanism enables robust zero-shot forecasting across diverse datasets \cite{woo_unified_2024}. Chronos adapts language-modeling techniques by tokenizing time series data for transformer input, enabling probabilistic forecasting with strong zero-shot performance across multiple datasets \cite{ansari2024chronos}. Lastly, TimesFM employs a decoder-only architecture with input patching, achieving promising zero-shot results on real-world and synthetic datasets while bridging classical statistical methods and deep learning \cite{das_decoder-only_2024}.
So, while advanced models have been introduced, their applicability for gap-filling in smart meter data remains underexplored, motivating this study.

\section{Research approach}
\label{sec:03_research_approach}
To benchmark the performance of various models—spanning statistical, \gls{ml}, and \glspl{llm} within the energy sector, we followed this research process. First, we selected a dataset and prepared the data for analysis. Next, we trained \gls{ml} models and performed inference using the selected \gls{tsfm}. Finally, we evaluated their performance using a set of standard error metrics to ensure a comprehensive comparison.

\subsection{Data Preparation}

We used a publicly available dataset containing household energy consumption data from London, which was collected through smart meters in 2013 \cite{uk_power_networks_smartmeter_2014}. It includes energy consumption measurements (kWh) recorded at half-hour intervals over a year for 5567 residential consumers.

Since the dataset is publicly accessible, available \glspl{llm} might have used it during their training. To address this potential concern, we applied an anonymization technique following \cite{domingo2005ordinal}. The anonymization ensured that any model evaluations would reflect genuine predictive performance, rather than potential recall of previously seen data.

\begin{figure}[h!]
    \centering
    \includegraphics[width=\columnwidth]{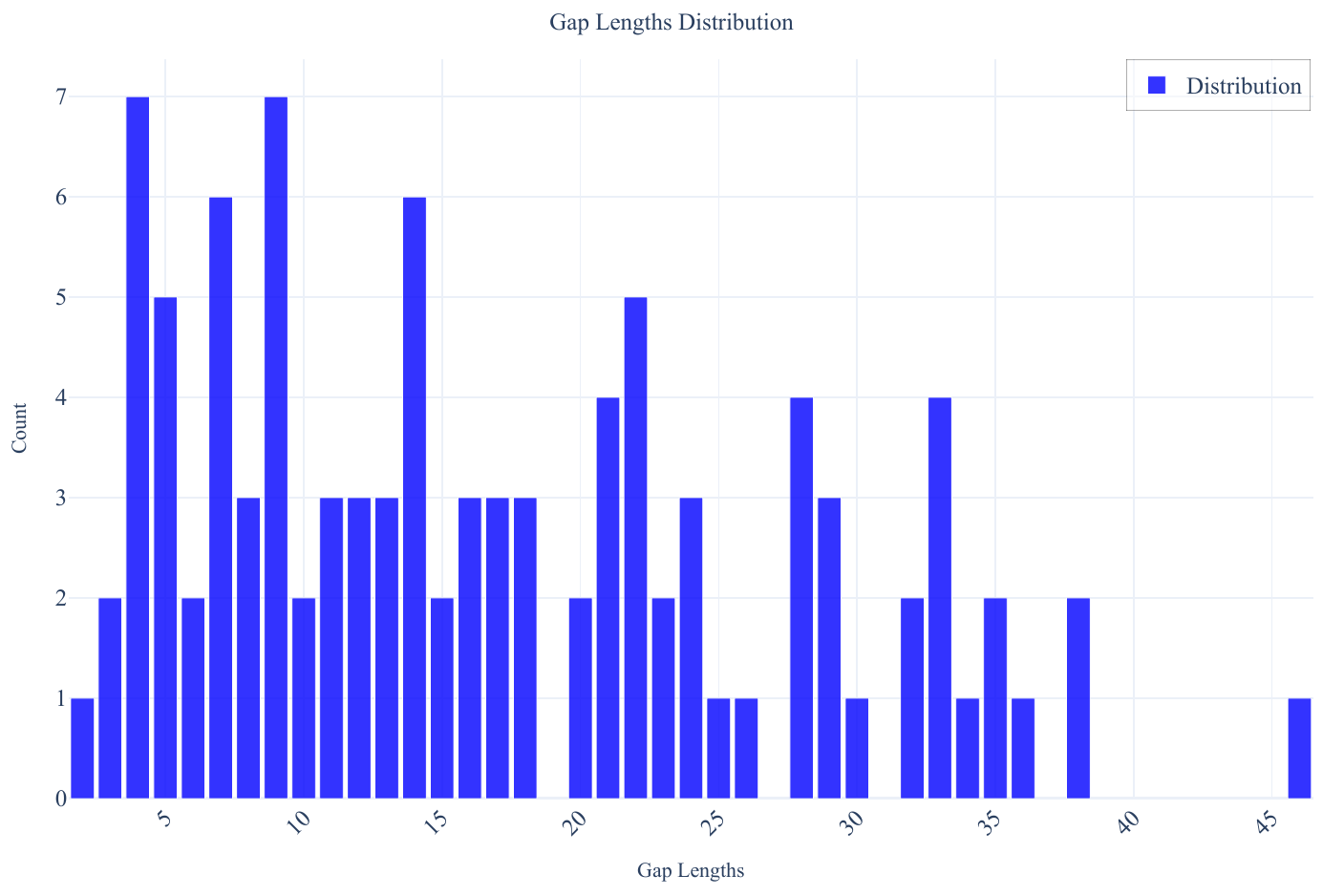}
    \caption{Distribution of randomly generated gaps across the evaluated data points}
    \label{fig:gapdistro}
\end{figure}

After anonymization, we randomly selected 10 smart meters from the dataset. For each meter, we created 10 random gaps using a uniform random distribution. The gap sizes varied, with a maximum size of up to 48 entries, corresponding to a full day of missing data. Figure \ref{fig:gapdistro} illustrates the distribution of these gaps, demonstrating that a range of gap lengths was considered in our experiments. The overall dataset spans a one-year period.

\subsection{Experiments}
After preparing the data, we selected a range of models across academic and industrial literature for benchmarking. To serve as baseline systems, we included four basic models: \gls{slp}, Padded Last, Last Week, and Linear Interpolation. These models were chosen because they represent some of the earliest and simplest approaches used for time series forecasting.

For statistical and \gls{ml} models, we focused on widely-used methods in state-of-the-art research. We selected five statistical models: \gls{arima}, Holt-Winters, Kalman Smoothing, Seasonal Naive, and \gls{mstl}. Additionally, we included four \gls{ml} models: \gls{xgb}, Random Forest, \gls{lgbm}, and \gls{knn}.

In the category of \glspl{llm}, we included two general-purpose language models: GPT-4o (commercial, closed-source) and Llama 3.1 405B (open-source). Additionally, we incorporated five top-performing \glspl{tsfm} specifically pre-trained for time series data. Among these, four are publicly available: TimesFM from Google, Moirai-1.1-R-large, Time-MoE, and Chronos-T5 (Large) from Amazon. We also evaluated TimeGPT, a commercial, closed-source model specifically trained on time series data.

\begin{table}[b!]
\centering
\centering
\caption{Comparison of forecasting models using five evaluation metrics. Bold green underlined values indicate the overall best-performing model, while bold blue underlined values show the overall least-performing model. Category-specific best and least performances are marked in green and blue, respectively.}
\label{tab:results}
\resizebox{\textwidth}{!}{%
\begin{NiceTabular}{>{\centering\arraybackslash}p{0.2\textwidth}p{0.15\textwidth}cccccc}[]
\toprule
\textbf{Category} & \textbf{Model} & \textbf{MAE} & \textbf{MAPE} & \textbf{MSE} & \textbf{RMSE} & \textbf{SMAPE} & \textbf{Ref} \\ 
\midrule
\midrule

\rowcolor{gray!10} \Block{4-1}{}{Baseline} & \gls{slp} & 
  {\color[HTML]{3531FF} {\ul \textbf{0.219 (0.0476)}}} &
  {\color[HTML]{3531FF} {\ul \textbf{1.2221 (0.583)}}} &
  {\color[HTML]{3531FF} {\ul \textbf{0.0678 (0.0235)}}} &
  {\color[HTML]{3531FF} {\ul \textbf{0.2566 (0.0459)}}} &
  {\color[HTML]{3531FF} {\ul \textbf{0.323 (0.0937)}}} &
 \cite{domingo2005ordinal} \\
 
\rowcolor{white} & Last Week &
  0.1475 (0.052) &
  0.6708 (0.1816) &
  0.0432 (0.0263) &
  0.1998 (0.0607) &
  0.254 (0.0543) & \cite{harman_systematic_nodate} \\
 
  \rowcolor{gray!10} & Padded Last &
  0.1066 (0.0316) &
  {\color[HTML]{036400} 0.3757 (0.0545)} &
  0.0263 (0.0146) &
  0.1561 (0.0459) &
  0.1985 (0.0432) &
\cite{harman_systematic_nodate} \\
  \rowcolor{white} & Linear Interpolation &
  {\color[HTML]{036400} 0.0961 (0.0325)} &
  0.3966 (0.1163) &
  {\color[HTML]{036400} 0.0208 (0.0132)} &
  {\color[HTML]{036400} 0.138 (0.0445)} &
  {\color[HTML]{036400} 0.1694 (0.025)} & \cite{noauthor_linear_nodate} \\
\midrule
\rowcolor{gray!10} \Block{5-1}{}{Statistical} & \gls{arima} &
  {\color[HTML]{3531FF} 0.0985 (0.0401)} &
  {\color[HTML]{3531FF} 0.4709 (0.1724)} &
  0.0177 (0.0139) &
  0.126 (0.0454) &
  {\color[HTML]{3531FF} 0.1831 (0.0453)} &
  \cite{box_box_2013}
   \\
  \rowcolor{white} & Kalman Smoothing &
  0.0955 (0.0315) &
  0.3987 (0.1067) &
  {\color[HTML]{3531FF} 0.0203 (0.0131)} &
  {\color[HTML]{3531FF} 0.136 (0.0448)} &
  0.1688 (0.0233) &
  \cite{kalman1960new}
   \\

  \rowcolor{gray!10} & Seasonal Naive &
  0.0861 (0.0292) &
  0.3367 (0.0542) &
  0.0162 (0.0096) &
  0.1225 (0.037) &
  0.1502 (0.0169) &
  \cite{mclaughlin1983forecasting}
   \\

  \rowcolor{white} & \gls{mstl} &
  0.0855 (0.0279) &
  0.3413 (0.0827) &
  0.0152 (0.0084) &
  0.119 (0.0338) &
  0.1566 (0.0262) &
  \cite{bandara_mstl_2022}
   \\
\rowcolor{gray!10} &
  Holt Winters &
  {\color[HTML]{036400} 0.0722 (0.0253)} &
  {\color[HTML]{036400} 0.2981 (0.0631)} &
  {\color[HTML]{036400} 0.0105 (0.0071)} &
  {\color[HTML]{036400} 0.0982 (0.0311)} &
  {\color[HTML]{036400} 0.1352 (0.0231)} &
  \cite{chatfield_holt-winters_1978}
   \\
\midrule
\rowcolor{white} \Block{4-1}{}
  {\gls{ml}} &
  XGB &
  {\color[HTML]{3531FF} 0.0936 (0.0339)} &
  {\color[HTML]{3531FF} 0.3984 (0.1155)} &
  {\color[HTML]{3531FF} 0.0179 (0.0125)} &
  {\color[HTML]{3531FF} 0.1271 (0.0437)} &
  {\color[HTML]{3531FF} 0.1787 (0.0378)} &
  \cite{chen_xgboost_2016}
   \\
  \rowcolor{gray!10} &
  \gls{lgbm} &
  0.0883 (0.0303) &
  0.3714 (0.0874) &
  0.0152 (0.0091) &
  0.1183 (0.0364) &
  0.1683 (0.0284) &
  \cite{ke_lightgbm_2017}
   \\
  \rowcolor{white} &  \gls{knn} &
  0.089 (0.0299) &
  0.3648 (0.0595) &
  0.0161 (0.0095) &
  0.122 (0.0371) &
  0.1681 (0.0348) &
  \cite{fix_discriminatory_1989}
   \\
  \rowcolor{gray!10} &
  Random Forest &
  {\color[HTML]{009901} 0.0861 (0.0313)} &
  {\color[HTML]{009901} 0.3636 (0.0995)} &
  {\color[HTML]{009901} 0.0149 (0.0098)} &
  {\color[HTML]{009901} 0.1164 (0.039)} &
  {\color[HTML]{009901} 0.1604 (0.0298)} &
  \cite{Breiman2001}
   \\
 \midrule
\rowcolor{white} \Block{7-1}{}{\gls{llm} \& \gls{tsfm} } &
  Llama 3.1 405B &
  {\color[HTML]{3531FF} 0.1083 (0.0473)} &
  {\color[HTML]{3531FF} 0.4594 (0.1848)} &
  {\color[HTML]{3531FF} 0.0263 (0.0264)} &
  0.1499 (0.0649) &
  {\color[HTML]{3531FF} 0.1932 (0.0497)} &
  \cite{metaIntroducingLlama}
     \\
  \rowcolor{gray!10} & GPT-4o &
  0.1063 (0.0336) &
  0.4232 (0.1044) &
  0.0246 (0.0152) &
  {\color[HTML]{3531FF} 0.151 (0.0447)} &
  0.1923 (0.0356) & \cite{achiam2023gpt}
   \\ 
  \rowcolor{white} & TimeGPT &
  0.0986 (0.0389) &
  0.4444 (0.1443) &
  0.0183 (0.0144) &
  0.128 (0.046) &
  0.181 (0.0408) &
  \cite{noauthor_nixtlanixtla_nodate}
   \\ 
  \rowcolor{gray!10} & Moirai-1.1-R-large &
  0.0739 (0.0204) &
  0.2856 (0.0532) &
  0.0118 (0.0052) &
  0.1065 (0.0229) &
  0.1364 (0.0196) &
  \cite{woo_unified_2024}
   \\
  \rowcolor{white} & TimesFM &
  0.0768 (0.025) &
  0.3343 (0.0888) &
  0.0116 (0.0064) &
  0.1043 (0.0281) &
  0.1443 (0.0262) &
  \cite{das_decoder-only_2024}
   \\
  \rowcolor{gray!10} & Chronos-T5 (Large) &
  0.0738 (0.0227) &
  {\color[HTML]{009901} {\ul \textbf{0.2584 (0.0416)}}} &
  0.0129 (0.0064) &
  0.1099 (0.0293) &
  0.1354 (0.02) &
  \cite{ansari2024chronos}
   \\ 
  \rowcolor{white} & Time-MoE &
  {\color[HTML]{009901} {\ul \textbf{0.0703 (0.023)}}} &
  0.2962 (0.0664) &
  {\color[HTML]{009901} {\ul \textbf{0.0098 (0.0056)}}} &
  {\color[HTML]{009901} {\ul \textbf{0.0957 (0.0275)}}} &
  {\color[HTML]{009901} {\ul \textbf{0.1313 (0.0218)}}} &
  \cite{shi_time-moe_2024} \\
\bottomrule
\end{NiceTabular}%
}
\end{table}

\begin{figure}[h!]
    \centering
    \includegraphics[width=0.7\columnwidth]{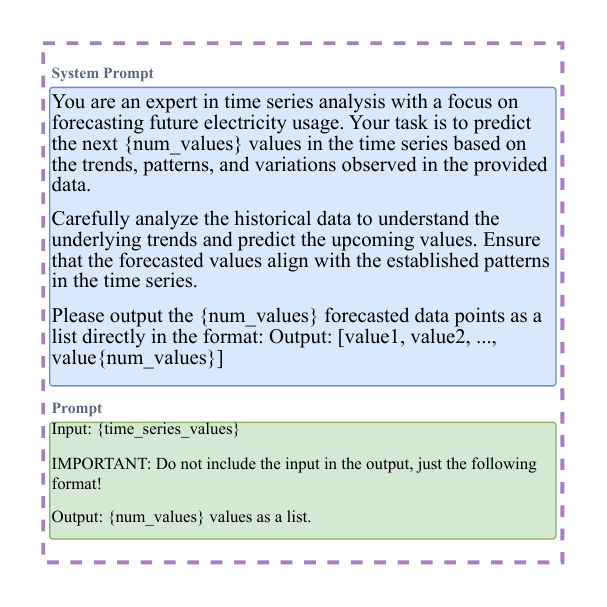}
    \caption{Prompts used for GPT-4o and Llama 3.1}
    \label{fig:prompts}
\end{figure}
We leveraged a diverse selection of models from different categories, enabling a well-rounded comparison and evaluation of their performance in the data imputation task by transforming the problem into a forecasting challenge. Our approach involved three key steps: \gls{fp}, \gls{bp}, and finally \gls{I}.

\begin{enumerate}
    \item \textbf{\gls{fp} (Forward Prediction):} We selected historical data from the left-hand side of a gap in the time series. The length of this historical data was seven days before the gap. Using this historical data, we forecasted values over a horizon equal to the length of the gap, which is referred to as the \gls{fp}.
    
    \item \textbf{\gls{bp} (Backward Prediction):} Similarly, we used historical data from the right-hand side of the gap to generate forecasts over the same horizon length, known as the \gls{bp}.
    
    \item \textbf{\gls{I} (Interpolation):} Once we obtained both forward and backward predictions, we interpolated between these predictions to generate the final imputed values. The interpolation was performed using the following formula:

    \begin{equation}
    \label{eq:interpolation}
    \text{\gls{I}}[i] = \frac{(\text{\gls{bp}}[i]^R \cdot i) + (\text{\gls{fp}}[i] \cdot (L -1 - i))}{L-1},
    \end{equation}

    where:
    \begin{itemize}
        \item \( i \) is the index within the gap length \( L \).
        \item \(\text{\gls{bp}}[i]^R\) is the \(i\)-th value of the \gls{bp} (reverted to align with the gap).
        \item \(\text{\gls{fp}}[i]\) is the \(i\)-th value of the \gls{fp}.
        \item \( L \) is the total length of the gap.
    \end{itemize}
    Equation \ref{eq:interpolation} calculates a weighted average of the forward and backward predictions, with weights decreasing linearly from the \gls{fp} to the \gls{bp} as \( i \) increases. This results in a smooth transition between the predictions, effectively filling the gap with an interpolated value that balances information from both sides of the gap.
    
\end{enumerate}

We employed the Nixtla stack \cite{olivares2022library_neuralforecast,garza2022statsforecast} and scikit-learn \cite{pedregosa_scikit-learn_2011} to train the statistical and \gls{ml} models. For the \gls{llm} models, we utilized APIs from ThebAI \cite{limited_terms_2023} and Nixtla \cite{noauthor_nixtlanixtla_nodate}, specifically accessing GPT-4 and Llama 3.1 through ThebAI, and the TimeGPT model via Nixtla. Figure \ref{fig:prompts} illustrates the prompts used for GPT-4 and Llama 3.1 in this process. The remaining time series forecasting models were executed on an Apple M3 Pro with 18 GB of memory. The cost for using API-based models amounted to approximately 100 Euros for ThebAI services and around 200 Euros for TimeGPT at the time of writing this manuscript.

\subsection{Evaluation}
We evaluated model performance using five standard error metrics: \gls{mae}, \gls{mse}, \gls{smape}, \gls{rmse}, and \gls{mape}. To account for model variability, each metric was computed based on the average of five runs. The error values were first averaged per household and then aggregated across all households. Additionally, we calculated the \gls{sd} for each metric to provide further insights into model stability and consistency.


\section{Results}
\label{sec:05_results}

Table \ref{tab:results} summarizes the performance of the selected models using five evaluation metrics: \gls{mae}, \gls{mape}, \gls{mse}, \gls{rmse}, and \gls{smape}. Each cell contains two values: the primary value representing the error metric and a secondary value, presented in parentheses, indicating the corresponding \gls{sd}. The models are categorized into four groups for comparison: Baseline, Statistical, \gls{ml}, and \glspl{llm}.

In this study, we used Baseline models as reference methods to establish a fundamental level of data imputation performance for the household electricity consumption dataset. These models rely on simple assumptions and easily accessible data, providing a straightforward benchmark for comparison against more sophisticated techniques.

Among these models, the \gls{slp} employs a basic extrapolation approach, predicting missing values based solely on preceding data points. This simplicity, however, led to the weakest performance, as \gls{slp} struggled to capture the complex patterns inherent in the dataset. In contrast, the Linear Interpolation method proved to be the most effective within the baseline category. By estimating missing values using surrounding data points, it consistently demonstrated lower errors and higher accuracy across all evaluation metrics. While these baseline models are inherently limited, they emphasize the challenge posed by data gaps and underscore the necessity of more advanced imputation methods.

The Statistical models analyzed in this study utilize established mathematical and probabilistic techniques to address time-series imputation. Among these methods, Holt Winters demonstrated strong performance, consistently achieving lower error rates across all metrics, making it the most effective statistical model in this category. Its ability to capture both seasonality and trends enabled it to adapt well to the complex patterns within the dataset. Conversely, ARIMA showed the weakest performance in this category, ranking lowest across three of the five evaluation metrics. Kalman Smoothing also faced notable challenges, underperming in two of the five metrics.


The evaluation of \gls{ml} models, including Random Forest, \gls{lgbm}, \gls{knn}, and \gls{xgb}, revealed differences in their performance. Random Forest emerged as the strongest model, achieving consistently lower error rates across all metrics. In contrast, \gls{xgb} recorded the highest error rates, marking it as the weakest performer within this group.

The evaluation of \glspl{llm} revealed a diverse range of performance levels across the metrics. Time-MoE emerged as the top performer overall, achieving the lowest error rates across most metrics, with the exception of \gls{mape}, where Chronos-T5 (Large) demonstrated superior results.

In contrast, Llama 3.1 405B had the weakest performance, recording the highest error rates for four out of five metrics. GPT-4o also displayed relatively poorer results in terms of \gls{rmse}. This behavior from Llama and GPT models aligns with expectations, as they are general-purpose models not specifically optimized for time-series tasks.



\section{Discussion}
\label{sec:06_discussion}
\begin{figure}[h!]
    \centering    \includegraphics[width=\columnwidth]{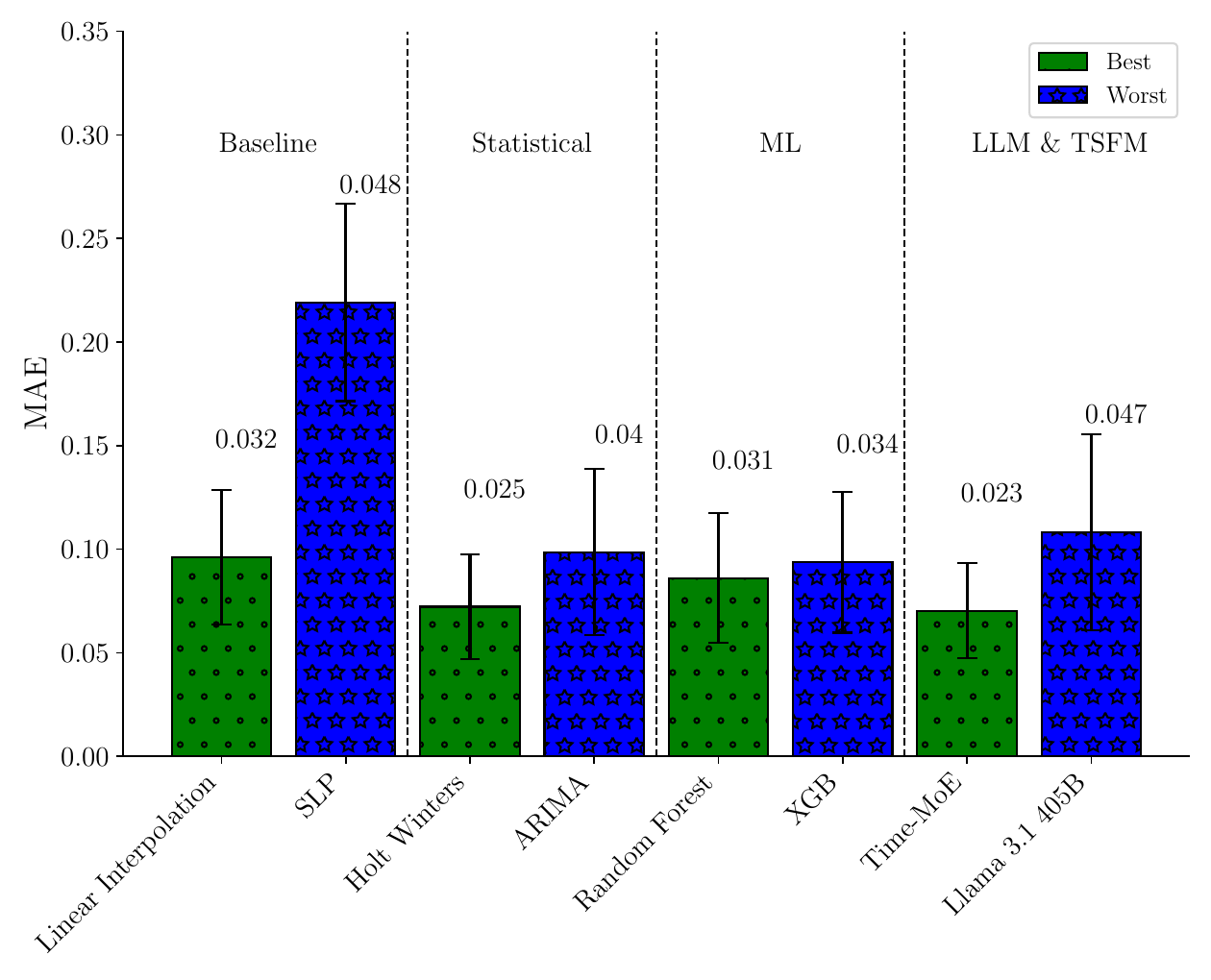} 
    \caption{\gls{mae} and \gls{sd} of the best-performing models (green) and least-performing models (blue) per category}
    \label{fig:maesd}
\end{figure}
\begin{figure}[h!]
    \centering    \includegraphics[width=\columnwidth]{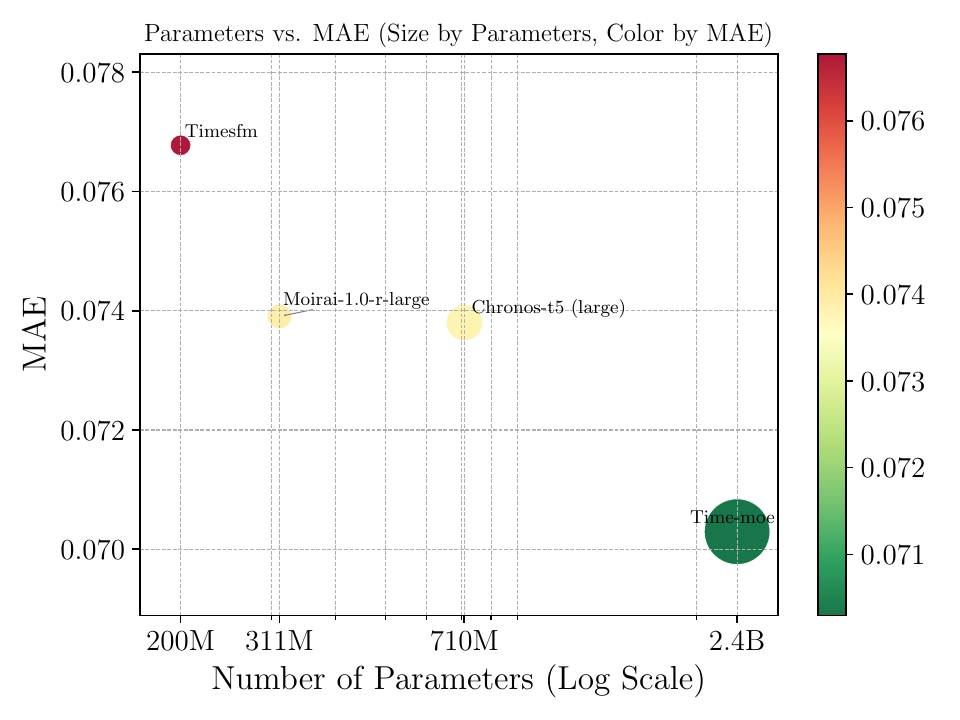}
    \caption{Relationship Between Parameters and \gls{mae}: Bubble Size Represents Parameter Count, Color Indicates \gls{mae} (Lower is Better)}
    \label{fig:maeparametrs}
\end{figure}
\begin{figure}[]
    \centering
    \includegraphics[width=\columnwidth]{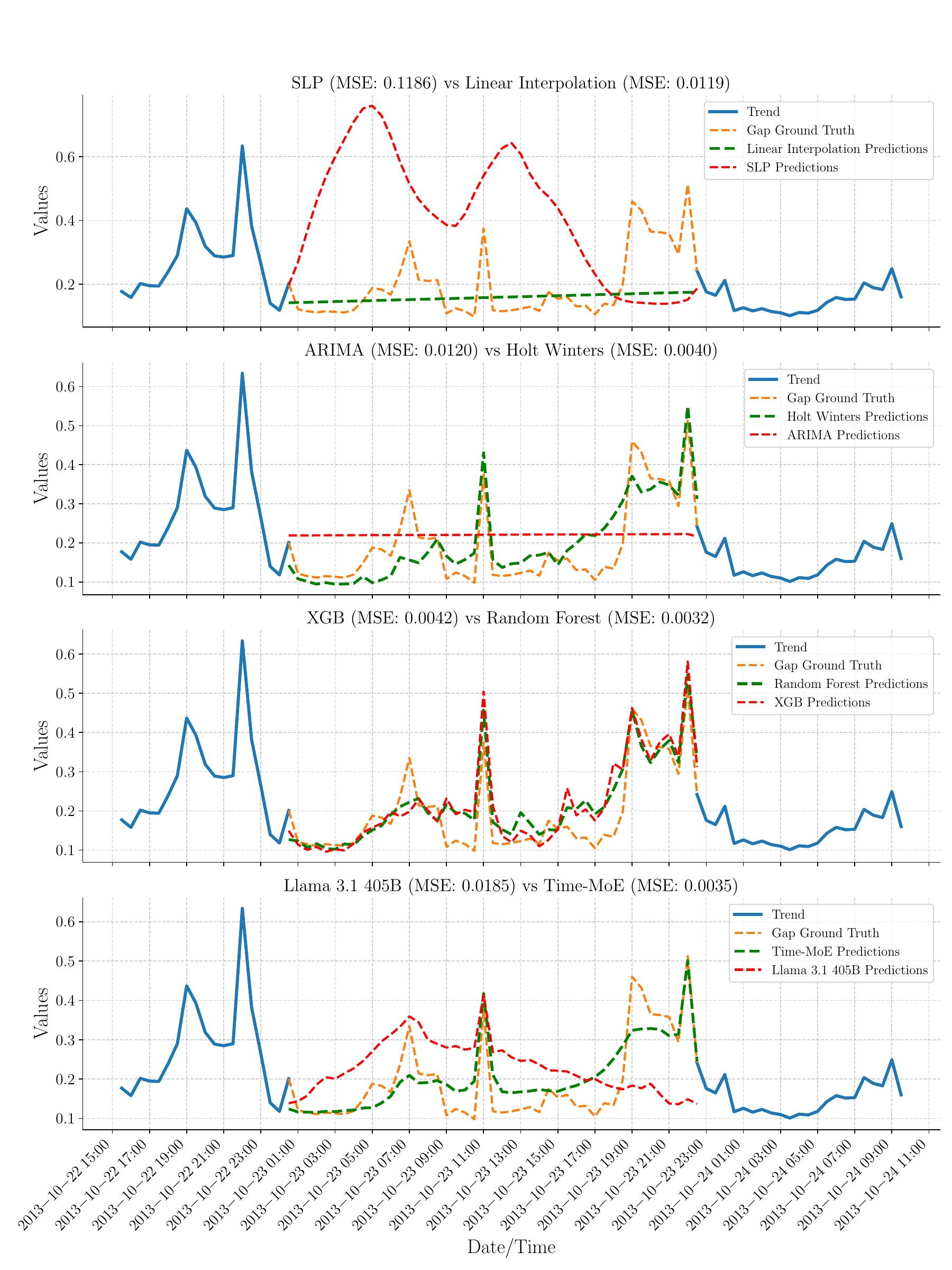} 
    \caption{Gap-filling results for the best and least-performing models on a household electricity consumption time series}
    \label{fig:forecastcomparison}
\end{figure}

Pre-trained time series forecasting models typically contain a significantly higher number of parameters compared to traditional machine learning models, resulting in a greater demand for computational power. However, their extensive training across large datasets allows them to perform forecasting tasks even in scenarios where limited data is available. In this study, we evaluated these models’ inference performance without fine-tuning on the target data. Our results indicate that, despite the lack of customization, these models often outperform machine learning models trained directly on similar data distributions.

Additionally, the training of traditional machine learning models often requires expertise to select features, optimize parameters, and tune hyperparameters. In contrast, pre-trained models, having been optimized on vast datasets, offer a more accessible and user-friendly solution for many applications, reducing the need for extensive domain expertise. This makes them a compelling choice for scenarios where ease of use, generalization, and scalability are prioritized.

Figure \ref{fig:maesd} presents the top- and lowest-performing models per category, evaluated using \gls{mae} and \gls{sd}.

The results reveal a clear relationship between lower error rates and smaller standard deviations, suggesting that models with superior predictive accuracy are also more stable and reliable. Notably, Time-MoE stands out by achieving the lowest \gls{mae} while maintaining relatively low variability, indicating consistent performance across different data samples.

The comparison across categories further underscores that statistical and \gls{ml} models, such as Holt-Winters and Random Forest, offer competitive results with moderate levels of variability. This suggests their effectiveness in capturing underlying trends and seasonality within time series data, even if they exhibit a slightly higher \gls{sd} compared to more advanced models.

In Figure \ref{fig:maeparametrs}, we present a comparison of the top four pre-trained time series forecasting models used in our experiments, focusing on their \gls{mae} relative to their number of parameters. The chart highlights that TimeMoE delivered strong performance across all models, despite having the highest number of parameters among the time series-specific models analyzed. Llama 3.1 and GPT-4, with significantly higher parameter counts (405 billion and over 200 billion, respectively), were excluded as they are not specialized for time series tasks. TimeGPT was also omitted due to a lack of available information on its parameter count.

Another notable observation is that TimesFM, developed by Google, demonstrated the lowest performance among these four models but still achieved competitive results, particularly given its comparatively lower parameter count. This suggests that, despite differences in architecture and complexity, trade-offs between model size and accuracy warrant careful consideration in resource-constrained settings.

In Figure \ref{fig:forecastcomparison}, we illustrate a short-term segment of household consumption data to compare the performance of the best and least effective models within our defined categories. From the visualizations, it is clear that simpler models like \gls{slp}, Linear Interpolation, and ARIMA struggle to capture the inherent patterns and spikes in the time series. Their predictions tend to be relatively flat and lack the ability to adapt to sudden changes or complex dynamics present in the data.

On the other hand, models such as Holt-Winters, Random Forest, \gls{xgb}, and TimeMoE exhibit a superior capacity to identify and follow spikes and fluctuations in the data. This demonstrates their ability to detect and adapt to intricate patterns within the time series, suggesting their greater potential for handling data with variability and noise.


\section{Conclusions and future work}
\label{sec:07_conclusion_and_future_work}
In this study, we evaluated the performance of traditional statistical methods, \gls{ml} models, \glspl{llm}, and \glspl{tsfm} for time-series data imputation, using the publicly available London smart meter household dataset. Our findings show that traditional statistical methods, such as Holt-Winters, and \gls{ml} models, such as Random Forest, performed well in capturing predictable trends, seasonal patterns, and addressing smart meter data gaps. These approaches demonstrated consistent and reliable results, emphasizing their viability for such tasks.

However, \glspl{tsfm}, particularly TimeMoE, outperformed all other approaches, demonstrating exceptional capabilities in handling data imputation even without specific training on the dataset—relying solely on inference. This suggests that \glspl{tsfm} bring unique advantages for filling data gaps, especially as smart grid data scales in complexity. As \glspl{tsfm} continue to evolve, they present new opportunities for enhancing predictive accuracy and data management in smart grids. Ultimately, the choice of approach should consider the specific requirements of the imputation task and available computational resources.

Future research should focus on fine-tuning open-source pre-trained time series models to optimize their performance for electricity consumption data. Additionally, experiments should be conducted with scenarios involving data gaps longer than one day to better identify models that perform well on extended gaps compared to those excelling with shorter gaps. Another valuable direction is to vary the length of historical data inputs to assess the effectiveness of inference-based predictions and to examine how fine-tuning strategies influence model adaptability. Collectively, these efforts will help identify optimal solutions for data imputation in smart grids, enhancing both predictive accuracy and reliability.


\bibliographystyle{ieeetr} 
\bibliography{template}

\begin{thebibliography}{10}

\bibitem{3_ds}
M.~L. {Di Silvestre}, S.~Favuzza, E.~{Riva Sanseverino}, and G.~Zizzo, ``How decarbonization, digitalization and decentralization are changing key power infrastructures,'' {\em Renewable and Sustainable Energy Reviews}, vol.~93, pp.~483--498, 2018.

\bibitem{power_systems_SOTA}
M.~{\c C}olak and E.~Irmak, ``A state-of-the-art review on electric power systems and digital transformation,'' {\em Electric Power Components and Systems}, vol.~51, no.~11, pp.~1089--1112, 2023.

\bibitem{2015_sm_taskforce}
{Smart Grids Task Force}, ``Interoperability, standards and functionalities applied in the large scale roll out of smart metering,'' tech. rep., Standards and Interoperability for Smart Grids Deployment (EG1) within the European Smart Grids Task Force, 2015.

\bibitem{2019_review_sm_analytics}
Y.~Wang, Q.~Chen, T.~Hong, and C.~Kang, ``Review of smart meter data analytics: Applications, methodologies, and challenges,'' {\em IEEE Transactions on Smart Grid}, vol.~10, no.~3, pp.~3125--3148, 2019.

\bibitem{7778826}
E.~Ebeid, R.~Heick, and R.~H. Jacobsen, ``Presenting user behavior from main meter data,'' in {\em 2016 IEEE International Conference on Smart Grid Communications (SmartGridComm)}, pp.~594--599, 2016.

\bibitem{Wu2022ReviewOS}
J.~Wu, A.~Koirala, and D.~V. Hertem, ``Review of statistics based coping mechanisms for smart meter missing data in distribution systems,'' {\em 2022 IEEE PES Innovative Smart Grid Technologies Conference Europe (ISGT-Europe)}, pp.~1--6, 2022.

\bibitem{jeong2021missing}
D.~Jeong, C.~Park, and Y.~M. Ko, ``Missing data imputation using mixture factor analysis for building electric load data,'' {\em Applied Energy}, vol.~304, p.~117655, 2021.

\bibitem{tang2024time}
H.~Tang, C.~Zhang, M.~Jin, Q.~Yu, Z.~Wang, X.~Jin, Y.~Zhang, and M.~Du, ``Time series forecasting with llms: Understanding and enhancing model capabilities,'' {\em arXiv preprint arXiv:2402.10835}, 2024.

\bibitem{wang_deep_2024}
J.~Wang, W.~Du, W.~Cao, K.~Zhang, W.~Wang, Y.~Liang, and Q.~Wen, ``Deep {Learning} for {Multivariate} {Time} {Series} {Imputation}: {A} {Survey},'' Feb. 2024.
\newblock arXiv:2402.04059 [cs].

\bibitem{pei_cross-dimensional_2024}
J.~Pei, J.~Ma, K.~L. Man, C.~Zhao, and Z.~Tian, ``A {Cross}-{Dimensional} {Attention} {Discriminating} {Masked} {Method} for {Building} {Energy} {Time}-{Series} {Data} {Imputation},'' in {\em 2024 9th {International} {Conference} on {Smart} and {Sustainable} {Technologies} ({SpliTech})}, (Bol and Split, Croatia), pp.~1--6, IEEE, June 2024.

\bibitem{hwang_cc-gain_2024-1}
J.~Hwang and D.~Suh, ``{CC}-{GAIN}: {Clustering} and classification-based generative adversarial imputation network for missing electricity consumption data imputation,'' {\em Expert Systems with Applications}, vol.~255, p.~124507, Dec. 2024.

\bibitem{vasenin_incorporating_2024-1}
D.~Vasenin, M.~Pasetti, D.~Astolfi, N.~Savvin, S.~Rinaldi, and A.~Berizzi, ``Incorporating {Seasonal} {Features} in {Data} {Imputation} {Methods} for {Power} {Demand} {Time} {Series},'' {\em IEEE Access}, vol.~12, pp.~103520--103536, 2024.
\newblock Conference Name: IEEE Access.

\bibitem{wang2024tssurvey}
Y.~Wang, H.~Wu, J.~Dong, Y.~Liu, M.~Long, and J.~Wang, ``Deep time series models: A comprehensive survey and benchmark,'' 2024.

\bibitem{lee2024evaluating}
K.~Lee, H.~Lim, J.~Hwang, and D.~Lee, ``Evaluating missing data handling methods for developing building energy benchmarking models,'' {\em Energy}, vol.~308, p.~132979, 2024.

\bibitem{du_tsi-bench_2024}
W.~Du, J.~Wang, L.~Qian, Y.~Yang, F.~Liu, Z.~Wang, Z.~Ibrahim, H.~Liu, Z.~Zhao, Y.~Zhou, W.~Wang, K.~Ding, Y.~Liang, B.~A. Prakash, and Q.~Wen, ``{TSI}-{Bench}: {Benchmarking} {Time} {Series} {Imputation},'' June 2024.
\newblock arXiv:2406.12747 [cs].

\bibitem{harman_systematic_nodate}
A.~Harman, L.~Baur, and A.~Sauer, ``Systematic comparison of imputation models for automatized gap filling on electrical load data of compressor composites in the industrial sector,'' {\em Energy Proceedings}, 2024.

\bibitem{meyer2024benchmarkingtimeseriesfoundation}
M.~Meyer, D.~Zapata, S.~Kaltenpoth, and O.~Müller, ``Benchmarking time series foundation models for short-term household electricity load forecasting,'' 2024.

\bibitem{tan2024language}
M.~Tan, M.~A. Merrill, V.~Gupta, T.~Althoff, and T.~Hartvigsen, ``Are language models actually useful for time series forecasting?,'' {\em arXiv preprint arXiv:2406.16964}, 2024.

\bibitem{garza2023timegpt1}
A.~Garza and M.~Mergenthaler-Canseco, ``Timegpt-1,'' 2023.

\bibitem{shi_time-moe_2024}
X.~Shi, S.~Wang, Y.~Nie, D.~Li, Z.~Ye, Q.~Wen, and M.~Jin, ``Time-{MoE}: {Billion}-{Scale} {Time} {Series} {Foundation} {Models} with {Mixture} of {Experts},'' Oct. 2024.
\newblock arXiv:2409.16040.

\bibitem{woo_unified_2024}
G.~Woo, C.~Liu, A.~Kumar, C.~Xiong, S.~Savarese, and D.~Sahoo, ``Unified {Training} of {Universal} {Time} {Series} {Forecasting} {Transformers},'' May 2024.
\newblock arXiv:2402.02592.

\bibitem{ansari2024chronos}
A.~F. Ansari, L.~Stella, C.~Turkmen, X.~Zhang, P.~Mercado, H.~Shen, O.~Shchur, S.~S. Rangapuram, S.~Pineda~Arango, S.~Kapoor, J.~Zschiegner, D.~C. Maddix, H.~Wang, M.~W. Mahoney, K.~Torkkola, A.~Gordon~Wilson, M.~Bohlke-Schneider, and Y.~Wang, ``Chronos: Learning the language of time series,'' {\em arXiv preprint arXiv:2403.07815}, 2024.

\bibitem{das_decoder-only_2024}
A.~Das, W.~Kong, R.~Sen, and Y.~Zhou, ``A decoder-only foundation model for time-series forecasting,'' Apr. 2024.
\newblock arXiv:2310.10688.

\bibitem{uk_power_networks_smartmeter_2014}
U.~P. Networks, ``{SmartMeter} {Energy} {Consumption} {Data} in {London} {Households},'' 2014.

\bibitem{domingo2005ordinal}
J.~Domingo-Ferrer and V.~Torra, ``Ordinal, continuous and heterogeneous k-anonymity through microaggregation,'' {\em Data Mining and Knowledge Discovery}, vol.~11, pp.~195--212, 2005.

\bibitem{noauthor_linear_nodate}
``Linear {Interpolation} - an overview {\textbar} {ScienceDirect} {Topics}.''

\bibitem{box_box_2013}
G.~Box, ``Box and {Jenkins}: {Time} {Series} {Analysis}, {Forecasting} and {Control},'' in {\em A {Very} {British} {Affair}: {Six} {Britons} and the {Development} of {Time} {Series} {Analysis} {During} the 20th {Century}} (T.~C. Mills, ed.), pp.~161--215, London: Palgrave Macmillan UK, 2013.

\bibitem{kalman1960new}
R.~E. Kalman, ``A new approach to linear filtering and prediction problems,'' {\em Transactions of the ASME--Journal of Basic Engineering}, vol.~82, no.~Series D, pp.~35--45, 1960.

\bibitem{mclaughlin1983forecasting}
R.~L. McLaughlin, ``Forecasting models: Sophisticated or naive?,'' {\em Journal of Forecasting (pre-1986)}, vol.~2, no.~3, p.~274, 1983.

\bibitem{bandara_mstl_2022}
K.~Bandara, R.~Hyndman, and C.~Bergmeir, ``{MSTL}: {A} {Seasonal}-{Trend} {Decomposition} {Algorithm} for {Time} {Series} with {Multiple} {Seasonal} {Patterns},'' {\em International Journal of Operational Research}, vol.~1, no.~1, p.~1, 2022.

\bibitem{chatfield_holt-winters_1978}
C.~Chatfield, ``The {Holt}-{Winters} {Forecasting} {Procedure},'' {\em Journal of the Royal Statistical Society. Series C (Applied Statistics)}, vol.~27, no.~3, pp.~264--279, 1978.
\newblock Publisher: [Royal Statistical Society, Oxford University Press].

\bibitem{chen_xgboost_2016}
T.~Chen and C.~Guestrin, ``{XGBoost}: {A} {Scalable} {Tree} {Boosting} {System},'' in {\em Proceedings of the 22nd {ACM} {SIGKDD} {International} {Conference} on {Knowledge} {Discovery} and {Data} {Mining}}, (San Francisco California USA), pp.~785--794, ACM, Aug. 2016.

\bibitem{ke_lightgbm_2017}
G.~Ke, Q.~Meng, T.~Finley, T.~Wang, W.~Chen, W.~Ma, Q.~Ye, and T.-Y. Liu, ``{LightGBM}: {A} {Highly} {Efficient} {Gradient} {Boosting} {Decision} {Tree},'' in {\em Advances in {Neural} {Information} {Processing} {Systems}}, vol.~30, Curran Associates, Inc., 2017.

\bibitem{fix_discriminatory_1989}
E.~Fix and J.~L. Hodges, ``Discriminatory {Analysis}. {Nonparametric} {Discrimination}: {Consistency} {Properties},'' {\em International Statistical Review / Revue Internationale de Statistique}, vol.~57, no.~3, pp.~238--247, 1989.
\newblock Publisher: [Wiley, International Statistical Institute (ISI)].

\bibitem{Breiman2001}
L.~Breiman, ``Random forests,'' {\em Machine Learning}, vol.~45, pp.~5--32, Oct 2001.

\bibitem{metaIntroducingLlama}
``{I}ntroducing {L}lama 3.1: {O}ur most capable models to date --- ai.meta.com.'' \url{https://ai.meta.com/blog/meta-llama-3-1/}.
\newblock [Accessed 14-10-2024].

\bibitem{achiam2023gpt}
J.~Achiam, S.~Adler, S.~Agarwal, L.~Ahmad, I.~Akkaya, F.~L. Aleman, D.~Almeida, J.~Altenschmidt, S.~Altman, S.~Anadkat, {\em et~al.}, ``Gpt-4 technical report,'' {\em arXiv preprint arXiv:2303.08774}, 2023.

\bibitem{noauthor_nixtlanixtla_nodate}
``Nixtla/nixtla: {TimeGPT}-1: production ready pre-trained {Time} {Series} {Foundation} {Model} for forecasting and anomaly detection. {Generative} pretrained transformer for time series trained on over {100B} data points..''

\bibitem{olivares2022library_neuralforecast}
K.~G. Olivares, C.~Challú, F.~Garza, M.~M. Canseco, and A.~Dubrawski, ``{NeuralForecast}: User friendly state-of-the-art neural forecasting models..'' {PyCon} Salt Lake City, Utah, US 2022, 2022.

\bibitem{garza2022statsforecast}
F.~Garza, M.~M. Canseco, and K.~G.~O. Cristian~Challú, ``{StatsForecast}: Lightning fast forecasting with statistical and econometric models.'' {PyCon} Salt Lake City, Utah, US 2022, 2022.

\bibitem{pedregosa_scikit-learn_2011}
F.~Pedregosa, G.~Varoquaux, A.~Gramfort, V.~Michel, B.~Thirion, O.~Grisel, M.~Blondel, P.~Prettenhofer, R.~Weiss, V.~Dubourg, J.~Vanderplas, A.~Passos, D.~Cournapeau, M.~Brucher, M.~Perrot, and E.~Duchesnay, ``Scikit-learn: {Machine} {Learning} in {Python},'' {\em Journal of Machine Learning Research}, vol.~12, no.~85, pp.~2825--2830, 2011.

\bibitem{limited_terms_2023}
{The BAI Limited}, ``{Terms of Service},'' 2023.

\end{thebibliography}







\end{document}